\documentclass[twoside, 11pt]{article}
\usepackage{jair, rawfonts}

\arxivheading{1}{2020}{xx-xx}{11/20}{x/xx}

\ShortHeadings{The \texttt{Learn-to-Race} Autonomous Racing Virtual Challenge}
{\footnotesize Francis, Chen, Ganju, Kathpal, Poonganam, Shivani, Vyas, Genc, Zhukov, Kumskoy, Koul, Oh, Nyberg}
\firstpageno{1}

\usepackage{amsfonts}
\usepackage{graphicx}
\usepackage[table]{xcolor} 
\usepackage{dsfont}
\usepackage{tabularx}
\usepackage{mathptmx}
\usepackage{pifont}
\usepackage{booktabs} 
\usepackage{comment}
\usepackage{natbib}
\usepackage[colorlinks=true,linkcolor=black,anchorcolor=black,citecolor=black,filecolor=black,menucolor=black,runcolor=black,urlcolor=blue]{hyperref}
\usepackage{bm}
\usepackage[parfill]{parskip}

\usepackage{amsmath}
\usepackage{amsfonts}

\usepackage{cleveref}
\usepackage{keyval}
\usepackage[ruled,vlined]{algorithm2e}
\usepackage{amssymb}
\usepackage{color}
\usepackage{wrapfig} 
\usepackage{url}

\newcommand{\parencite}[1]{\citep{#1}}%
\newcommand{\textcite}[1]{\cite{#1}}%
\newcommand{\para}[1]{{\noindent\textbf{#1}}}
\newcommand{\parai}[1]{{\noindent\textit{#1}}}
\newcommand{\cut}[1]{}

\newcommand{\ann}[1]{}

\newcommand{\LTR}{\texttt{Learn-to-Race}}
\newcommand{\ltr}{\texttt{L2R}}

\begin{document}

\raggedbottom 

\title{{Learn-to-Race} Challenge 2022: Benchmarking Safe Learning and Cross-domain Generalisation in Autonomous Racing}

\author{\name Jonathan Francis\thanks{Indicates equal contribution from these authors.}$\;\;$\thanks{Corresponding author | Jonathan Francis, jmf1@alumni.cmu.edu} \email jmf1@cs.cmu.edu \\
\addr School of Computer Science, Carnegie Mellon University, 5000 Forbes Ave., Pittsburgh, PA, USA \\
\addr Human-Machine Collaboration, Bosch Research Pittsburgh, 2555 Smallman St., Pittsburgh, PA, USA 
\AND
\name Bingqing Chen$^*$ \email bingqinc@andrew.cmu.edu \\
\addr College of Engineering, Carnegie Mellon University, 5000 Forbes Ave., Pittsburgh, PA, USA
\AND
\name Siddha Ganju$^*$ \email sganju@nvidia.com \\
\addr NVIDIA, 2788 San Tomas Expy, Santa Clara, CA, USA
\AND
\name Sidharth Kathpal$^*$ \email skathpal@cs.cmu.edu \\
\addr School of Computer Science, Carnegie Mellon University, 5000 Forbes Ave., Pittsburgh, PA, USA
\AND
\name Jyotish Poonganam$^*$ \email jyotish@aicrowd.com \\
\addr AICrowd, EPFL Innovation Park, Lausanne, CH
\AND
\name Ayush Shivani$^*$ \email ashivani@aicrowd.com \\
\addr AICrowd, EPFL Innovation Park, Lausanne, CH
\AND
\name Vrushank Vyas \email vrushank@aicrowd.com \\
\addr AICrowd, EPFL Innovation Park, Lausanne, CH
\AND
\name Sahika Genc \email sahika@amazon.com \\
\addr Amazon, 550 Terry Ave N, Seattle, WA
\AND
\name Ivan Zhukov \email zhukov@arrival.com \\
\addr Autonomous Driving \& Advanced Driver Assistance Systems, Arrival Ltd., London, UK W14 8TS
\AND
\name Max Kumskoy \email kumskoy@arrival.com \\
\addr Autonomous Driving \& Advanced Driver Assistance Systems, Arrival Ltd., London, UK W14 8TS
\AND
\name Anirudh Koul \email akoul@alumni.cmu.edu \\
\addr Pinterest, 651 Brannan St, San Francisco, CA, USA
\AND
\name Jean Oh \email jeanoh@cmu.edu \\
\addr School of Computer Science, Carnegie Mellon University, 5000 Forbes Ave., Pittsburgh, PA, USA
\AND
\name Eric Nyberg \email ehn@cs.cmu.edu \\
\addr School of Computer Science, Carnegie Mellon University, 5000 Forbes Ave., Pittsburgh, PA, USA}


\maketitle

\clearpage
\begin{abstract}

We present the results of our autonomous racing virtual challenge, based on the newly-released \LTR~(L2R) simulation framework, which seeks to encourage interdisciplinary research in autonomous driving and to help advance the state of the art on a realistic benchmark.
Analogous to racing being used to test cutting-edge vehicles, we envision autonomous racing to serve as a particularly challenging proving ground for autonomous agents as: (i) they need to make sub-second, safety-critical  decisions in a complex, fast-changing environment; and (ii) both perception and control must be robust to distribution shifts, novel road features, and unseen obstacles. 
Thus, the main goal of the challenge is to evaluate the joint safety, performance, and generalisation capabilities of reinforcement learning agents on multi-modal perception, through a two-stage process. In the first stage of the challenge, we evaluate an autonomous agent's ability to drive as fast as possible, while adhering to safety constraints. In the second stage, we additionally require the agent to adapt to an unseen racetrack through safe exploration. 
In this paper, we describe the new L2R Task 2.0 benchmark, with refined metrics and baseline approaches. We also provide an overview of deployment, evaluation, and rankings for the inaugural instance of the L2R Autonomous Racing Virtual Challenge (supported by Carnegie Mellon University, Arrival Ltd., AICrowd, Amazon Web Services, and Honda Research), which officially used the new \ltr~Task 2.0 benchmark and received over 20,100 views, 437 active participants, 46 teams, and 733 model submissions---from 88+ unique institutions, in 58+ different countries. Finally, we release leaderboard results from the challenge and provide description of the two top-ranking approaches in cross-domain model transfer, across multiple sensor configurations and simulated races.

\end{abstract}

\section{Introduction}
\label{sec:intro}

Autonomous driving has been gaining traction in the automotive and trucking industries, with the promise of enhancing safety, improving fuel efficiency, and reducing congestion \parencite{eugensson2013environmental, maurer2016autonomous}. Recently, notable failures in vehicle perception pipelines exposed severe \textit{limitations} in models' abilities to make safe, timely, and robust decisions, without compromising performance-oriented objectives. Several real-world autonomous racing challenges (e.g., \citet{roborace, indyac}) were introduced with the intention of deploying algorithms in \textit{Formula}-style track racing.  However, without high-fidelity racing simulation environment and interoperable training framework, experimentation of novel solutions in this space could prove costly and potentially dangerous.

Analogous to  racing being used as the proving ground for high-performance automotive technology, we also envision autonomous racing to serve as such for autonomous technology.  Autonomous agents are required to make sub-second, safety-critical decisions, in a complex, fast-changing environment; both perception and control must remain robust to distribution shifts, novel road features, and unseen obstacles, in order to facilitate cross-domain safety and performance. In \textcite{herman2021learn}, we released an open-source, high-fidelity simulation environment for high-speed racing, \LTR\footnote{See: \url{https://learn-to-race.org}}~(\ltr), with the hope to democratize autonomous racing research. In order to encourage use of \ltr~and to facilitate standardized benchmarking, we launch the \LTR~autonomous racing virtual challenge, with multi-institutional support from Carnegie Mellon University, Arrival Ltd., AICrowd, Amazon Web Services, and Honda Research. In its inaugural instance, the 5-month international virtual AI challenge enjoyed strong participation, with 20,100+ views, 437 individual participants, 46 teams, and 733 submissions---from 88+ unique institutions, in 58+ different countries.

\begin{figure}[t]
    \centering
    \includegraphics[width=\textwidth]{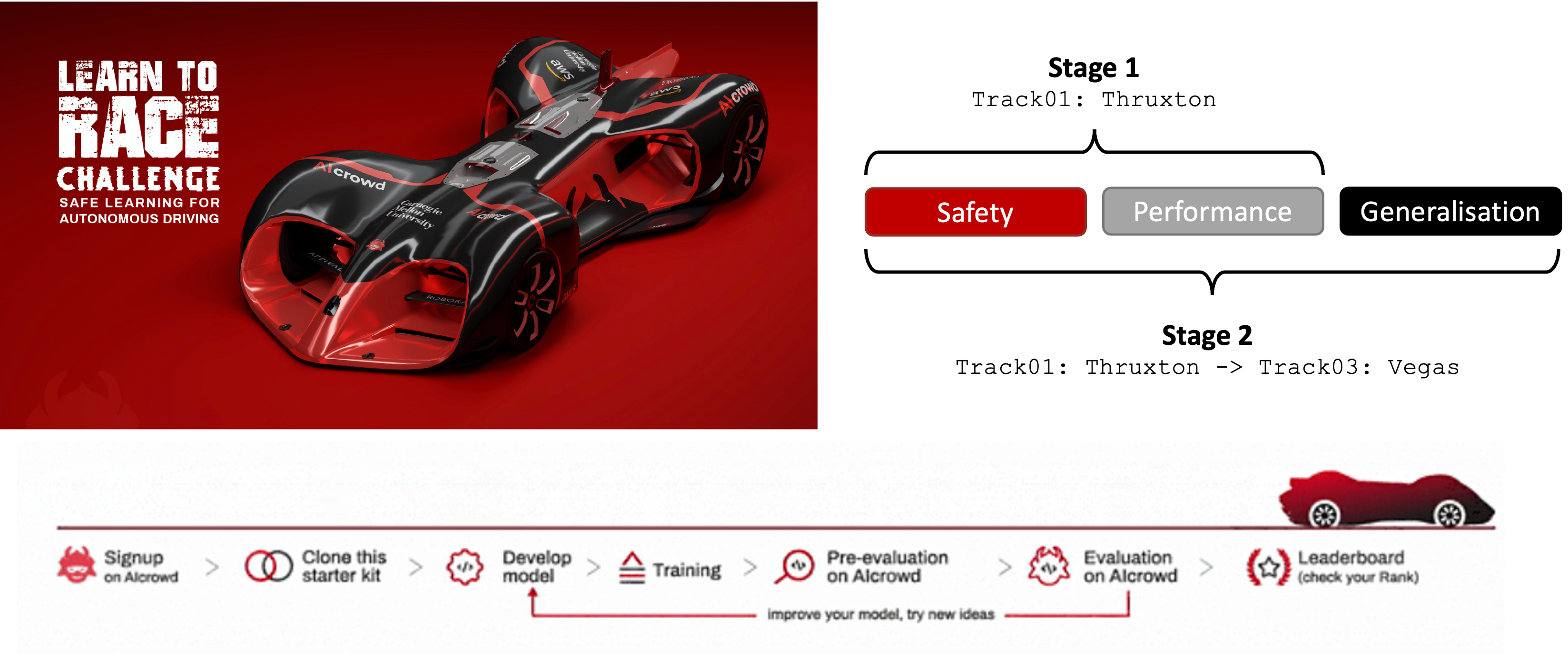}
    \caption{\LTR~Autonomous Racing Virtual Challenge overview.  The goal of the challenge is to evaluate the joint safety, performance, and generalisation
capabilities of reinforcement learning agents on multi-modal perception, through a two-stage process.}
    \label{fig:challenge}
\end{figure}

While the \ltr~framework was intended to facilitate a wide range of research topics, e.g., learning from demonstration, reinforcement learning (RL), model predictive control (MPC), sim-to-real transfer, and multimodal representation learning, we focus specifically on safe learning in the first instance of the \LTR~challenge. As autonomous technology advances, it is of paramount importance for autonomous vehicles to adheres to safety specifications, whether in urban driving or high-speed racing. Racing demands each vehicle to drive at its physical limits with little margin for safety, when any infraction could lead to catastrophic failures. Given this inherent tension, autonomous racing is a particularly challenging  task for safe learning algorithms.

Thus, the objective of the \LTR~challenge is to push the boundary of autonomous technology, with the goal of realizing the promise of improved safety in autonomous driving. In the challenge, participants develop RL agents to drive as fast as possible, while adhering to the safety constraints. Furthermore, participants are required to use high-dimensional visual data as inputs, in as opposed to low-dimensional features used in \cite{fuchs2020superhuman,wurman2022outracing}. Finally, we also test the agents' ability to adapt to a new environment, with a fixed time budget for safe exploration. Through this challenge, we pose the following fundamental research questions:

\begin{itemize}
\item How can an autonomous agent push performance towards its physical limits, while adhering to safety specifications?
\item How does an autonomous agent learn salient representations from high-dimensional sensory inputs that are generalisable and robust?
\item How should an autonomous agent explore safely and adapt to unseen scenarios?
\item How can we inject domain knowledge (e.g., model-based priors, common sense, logical rules, safety specifications, skill primitives, expert demonstrations) such that the autonomous agent is safer and more sample-efficient?
\item What models and architectures are realistic, to facilitate simulation-to-real transfer of models---to be integrated within autonomous racing vehicle software stacks?
\end{itemize}

In this paper, we release the \LTR~Task 2.0 benchmark, with refined metrics and baselines. We provide an overview of the inaugural \LTR~Autonomous Racing Virtual Challenge, including description of deployment, evaluation, and ranking procedures. Finally, we present the leaderboard results as well as descriptions of two top-ranking approaches, detailing strategies for cross-domain transfer, single- versus multi-camera configuration, and hybrid modeling.

The remainder of this manuscript is organized as follows: we provide high-level desiderata for autonomous racing vehicles in Section \ref{sec:desiderata}. In Section \ref{sec:review}, we summarize related work, existing simulation frameworks, and previous challenges in autonomous racing. In Sections \ref{sec:l2r} and \ref{sec:task2}, we briefly describe the \LTR~(\ltr) framework and release the official \ltr~Task 2.0 Benchmark. In Section \ref{sec:challenge}, we summarize the inaugural \ltr~autonomous racing virtual challenge and, in Section \ref{sec:results}, we provide the results of the challenge, in the form of leaderboard scores and top-ranking approach descriptions. We wrap up this manuscript with insights from the challenge and descriptions of future directions, in Section \ref{sec:future}, and with conclusions in Section \ref{sec:conclusions}. 

\section{Desiderata}\label{sec:desiderata}

In autonomous racing, vehicles must trade-off multiple (potentially-conflicting) objectives, in order to pursue safe, performant, and robust behaviour. Below, we outline some desired characteristics, as goals to achieve in this domain.

\begin{enumerate}
    \item \textit{Drives at high speeds}: as the primary metric for performance in autonomous racing, agents should drive as fast as possible and finish laps in the least amount of time.
    \item \textit{Does not incur in safety infractions}: in the pursuit of real-world deployment of autonomous racing algorithms, where system failures could have catastrophic ramifications, agents are not permitted to crash. Agents must also negotiate obstacles, turns, and other agents on the road, e.g., with an explicit model of safe behaviour.
    \item \textit{Adapts to unseen contexts}: the agent should be able to adapt and generalise to unseen contexts, such as new racetracks, new visual conditions (e.g., glare, shadows), and new dynamical regimes (e.g., effects of new weather conditions, different vehicles), that are not seen during training or initial periods of task-execution. At the limit, such an agent should be able to transfer from simulation to real-world environment. 
    \item \textit{Adheres to task structure}: regardless of whether the environment context is seen or unseen, the agent must appropriately adhere to task structure---e.g., staying on the drivable area, navigating turns and obstacles, and having an understanding of racing tactics and etiquette \citep{wurman2022outracing}.
    \item \textit{Balances multiple objectives}: In addition to driving as fast as possible while adhering to safety constraints,  the agent should simultaneously consider other objectives, such as  employ smooth and fuel-efficient maneuvers as much as possible, e.g., trading off cornering speed with boundary-relative position on the track.
    \item \textit{Algorithmic transferability}: the agents should be applicable to both simulation and the real-world environment. We envision Learn-to-Race framework will be interoperable with the real-world vehicle software stacks used by interfacing via ROS. 
\end{enumerate}

\section{Related Work}\label{sec:review}

In this section, we review prior work on autonomous racing, and existing simulators and similar competitions, which motivate and  inform the design of \LTR~challenge. 

\para{Autonomous racing.} One approach for autonomous racing is via MPC \citep{liniger2015optimization,rosolia2017autonomous,kabzan2019learning}, which solves a  planning problem iteratively over a receding time horizon with a model of the system dynamics. Aside from the challenges in modeling the complex dynamics, a significant drawback of such approach is the dependence on extensive sensor installation for localization and state estimation \citep{cai2021vision}. Another approach is to use a modular pipeline  \citep{kabzan2019learning, strobel2020accurate, francis2021core, tatiya2022knowledgedriven, betz2022autonomous}, starting from perception on raw sensory inputs, to localization and object-detection, and finally to planning and control. While this approach is most commonly used in practice, disadvantages of the approach include over-complexity and error propagation \parencite{yurtsever2020survey,francis2021core}. Recently, there is a lot of interest in using RL-based approaches for autonomous racing. In \citet{fuchs2020superhuman,chisari2021learning,wurman2022outracing}, RL agents were trained using low-dimensional features as inputs. In \citet{chen2015deepdriving,drews2017aggressive}, intermediate features were extracted from perception pipelines to determine control actions. In \citet{cai2021vision,weiss2020deepracing}, RL agents were trained end-to-end on visual inputs by imitating expert demonstration; in \citet{cai2021vision}, a data-driven model of the environment was further utilized to train the agent by unrolling future trajectories. 

While there is a large body of literature on end-to-end RL on perception data for urban driving \parencite{codevilla2018end,ohn2020learning,codevilla2019exploring,chen2020learning,zhang2021learning,prakash2021multi,zhang2021end,zhang2021learning}, there are significantly fewer works on the same topic for high-speed racing. We hypothesize this may partly be  attributed to  the lack of open-source, high-fidelity simulation environments for racing, in comparison to the ubiquity of CARLA simulator \cite{dosovitskiy2017carla} for urban driving research.  It is our hope that  \LTR~will democratize and facilitate autonomous racing research.


\para{Autonomous racing simulators and competitions.} \texttt{CarRacing-v0}, an OpenAI-gym environment \parencite{brockman2016openai}, is a simple racing environment, which provides a bird-eye view of 96$\times$96 pixels. TORCS \parencite{TORCS} is an open-source simulator, which enabled research on perception-based autonomous racing, such as \citet{chen2015deepdriving,drews2017aggressive}, and is also used by the Simulated Car Racing Championship \citep{loiacono2013simulated}. However, the vision quality of TORCS is low by contemporary standards and in comparison to the CARLA simulator for urban driving. Gran Turismo, a video game by Sony, has been used in for autonomous racing research in \citet{fuchs2020superhuman, wurman2022outracing}, wherein the researchers reported super-human performance from RL agents trained on a handful of selected features, e.g., vehicle states and way-points for upcoming track segments. However, Gran Turismo is not open-source, only runs on specialised gaming hardware, and provides no interface to control algorithms. These factors significantly limit the ability for external researchers to evaluate and benchmark on the environment. DeepRacer \parencite{balaji2020deepracer} is a open-source platform that support both simulation and real-world deployment of 1/18\textsuperscript{th} scale cars, which facilities research on end-to-end RL on raw visual inputs and sim-to-real transfer. There is also a recurring competition, DeepRacer League, based on the platform. F1TENTH \citep{o2020f1tenth} is a similar framework based on  1/10\textsuperscript{th} scale cars and has its competition series. In comparison to the aforementioned environments, \LTR~is a open-source simulation environment based on full-scale race cars with complex dynamics, and high-fidelity visual rendering.

Aside from DeepRacer and F1TENTH, which use reduced-scale cars, there are also a number of real-world challenges on full-scale race cars, such as RoboRace \citep{roborace}, Indy Autonomous Challenge \citep{indyac}, and AMZ Driverless \citep{AMZ}. Due to the significant cost for incurring safety infractions, the participants generally adopt a modular pipeline, instead of end-to-end RL. In comparison, a simulation environment allows for low-cost experimentation of novel algorithms. We envision the \LTR~framework, which closely follow the format of RoboRace, will enable sim-to-real transfer to RoboRace in the near future.

\section{\LTR~Framework}
\label{sec:l2r}
\begin{figure}
    \centering
    \includegraphics[width=0.292\textwidth]{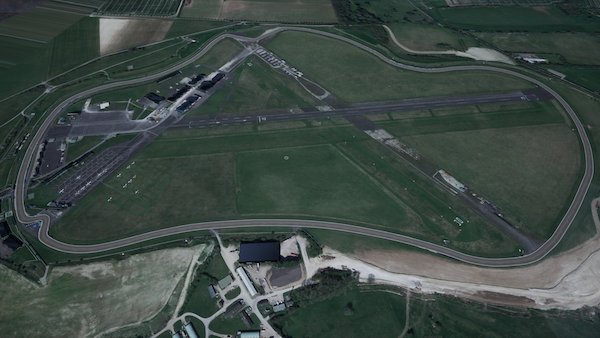}\hfill
    \includegraphics[width=0.3\textwidth]{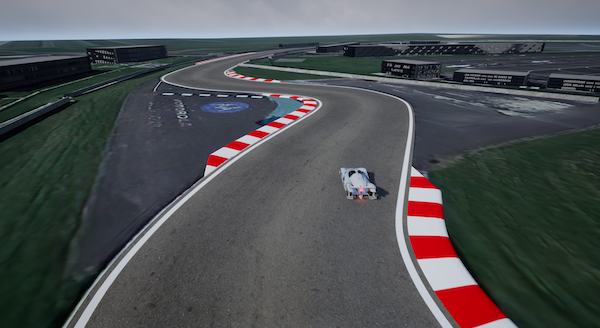}\hfill
    \includegraphics[width=0.3\textwidth]{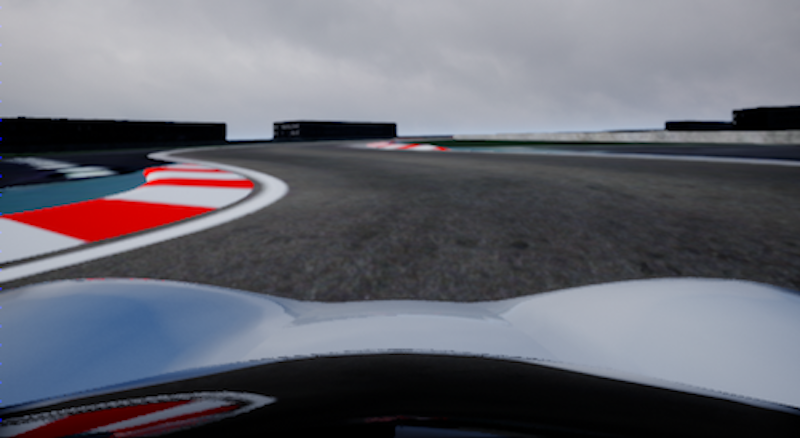}
    \\[\smallskipamount]
    \includegraphics[width=0.292\textwidth]{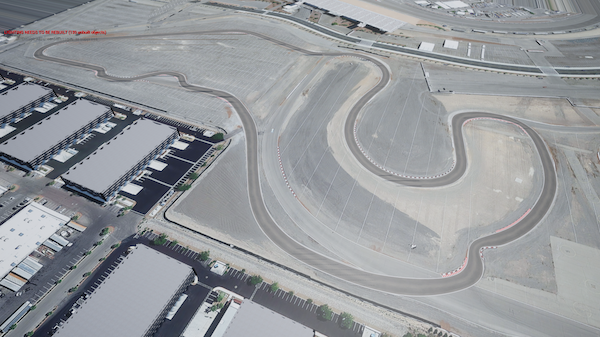}\hfill
    \includegraphics[width=0.3\textwidth]{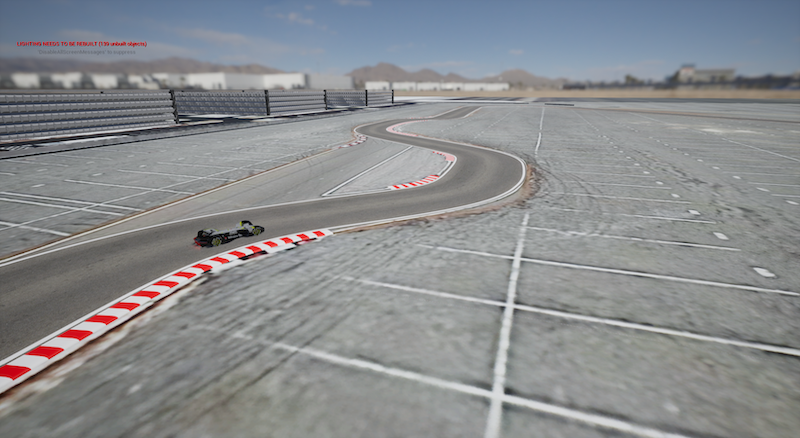}\hfill
       \includegraphics[width=0.3\textwidth]{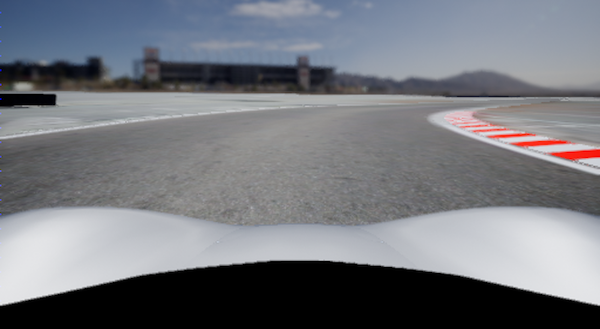}
    \\[\smallskipamount]
    \caption{\small\LTR~interfaces with a racing simulator, which features numerous real-world racetracks such as the Thruxton Circuit (\textit{top}) and Las Vegas Motor Speedway (\textit{bottom}) in bird-eye view (\textit{left}), third-person view (\textit{centre}), and ego-view (\textit{right}). }
    \label{fig:simulator_images}
\end{figure}

\begin{figure}
    \centering
    \includegraphics{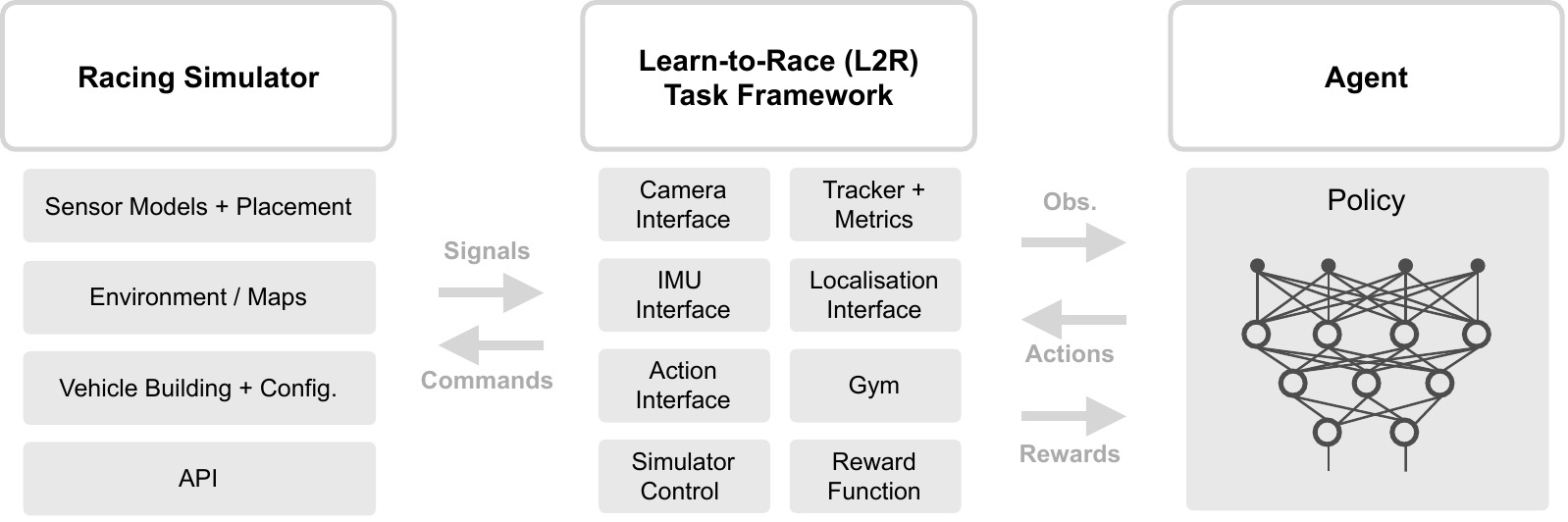}
    \caption{\LTR~Framwork}
    \label{fig:ltr}
\end{figure}

In our prior work \citep{herman2021learn}, we released \LTR~(\ltr): a python- and pytorch-based open-source, OpenAI Gym-compliant \parencite{brockman2016openai} framework which leverages a high-fidelity racing simulator or real-world vehicle interface (e.g., implemented in the Robot Operating System; ROS \citep{quigley2009ros}), for training and deployment of machine learning algorithms for continuous control contexts. We utilise the Arrival Ltd. autonomous racing simulator, which not only captures complex vehicle dynamics and renders photo-realistic views, but also plays a key role in bringing autonomous racing technology to real life---through official contribution to the Roborace Challenge series, the world’s first extreme competition of teams developing autonomous racing technologies. Figure \ref{fig:ltr} illustrates the \ltr~system components, and we refer interested readers to \citet{herman2021learn, chen2021safe} for further details about the framework.

\para{Learn-to-Race environment.} The \ltr~environment consists of four core component classes: \textsc{Controller}, \textsc{Environment}, \textsc{Agent}, and \textsc{Tracker}. The Controller initialises the backend (i.e., the simulator runtime or the ROS-based vehicle software stack) and initiates software communication interfaces with the backend's sensors, peripherals, and operation state-control mechanisms; after this initialisation step, the Controller can be used to perform `get' and `set' operations on the backend's configuration of sensor and vehicle parameters. The Environment class facilitates a training and deployment paradigm for machine learning models, in the form of popular OpenAI gyms, complete with the characterisation of the continuous control task in autonomous racing as a Markov decision process (MDP)---with an action space, an observation space, and reward/cost functions. The Environment's \texttt{make}, \texttt{step}, and \texttt{reset} methods, typical of MDP-inspired software control loops, are indeed implemented as adapters through the Controller's requisite interfaces to the backend. The Agent code implements a control policy, which maps \textit{observations} (received from the Environment, via the Controller's sensor interfaces with the backend) to \textit{actions} (sent to the Environment, for issuing vehicle control commands to the backend, via the Controller interfaces); prototypical Agent code is provided by the \ltr~framework, as templates for implementing the \texttt{select\_action} method, which is typical of approaches that operate within MDP-like training and inference settings. The Tracker module maintains a registry of the Agent's state, measures its progress towards task completion, records past success/failure conditions, and calculates the official task metrics.

\para{Autonomous racing simulator backend.} \ltr~supports the use of an autonomous driving simulator, as a backend API service, for developing policy algorithms for vehicle control. Currently, \ltr~uses the Arrival simulator \citep{herman2021learn}, which is a powerful tool for the development and testing of autonomous vehicles. It is based on Unreal Engine 4 and includes such features as: (i) a vehicle prototyping framework; (ii) full software-in-the-loop (SIL) simulation, to model all vehicle control devices; (iii) controller area network (CAN) bus interface; (iv) camera, inertial measurement unit (IMU), light detection and ranging (LiDAR), ultrasonic, and radar sensor models; (v) semantic segmentation; (vi) sensor placement and configuration facilities; (vii) V2V/V2I interface subsystem; (vii) dynamic racing scenario creation; (viii) race track generation from scanned datasets; (ix) support for full integration with the CARLA simulator \citep{dosovitskiy17carla}; and (x) an application programming interface (API), which is automatically generated based on C++ code analysis. The \LTR~framework maintains a connection with a simulator, via the interfaces established by the Controller, on initialisation and throughout task execution.


\section{\LTR~Task 2.0 Benchmark}
\label{sec:task2}

Alongside the \LTR~framework, we defined Task 1.0 in \citet{herman2021learn} as a benchmark for evaluating progress towards autonomous racing technology. In Task 2.0, we seek to assess agents on the basis of their joint safety, performance, and generalisability to unseen contexts. We describe the refined metrics and evaluation procedures, introduced in \ltr~Task 2.0.

\subsection{Primary Task Metrics}
\label{section:metrics}

We follow \citet{herman2021learn} in utilising our previous driving quality metrics, for the new benchmark: Episode Duration (ED), Average Adjusted Track Speed (AATS), Average Displacement Error (ADE), Trajectory Admissibility (TrA), Trajectory Efficiency (TrE), and Movement Smoothness (MS). Relative to these original Task 1.0 metrics, we also highlight the three important metrics for the Task 2.0 benchmark, introduced in this paper. In particular, we introduce success rate (as a proxy for safety), we retain average adjusted track speed (as a proxy for performance), and we introduce the total number of safety infractions.

\para{Success Rate.} Success rate (SR) is evaluated upon completion of a lap on a race track. Each race track is partitioned into a fixed number of segments, and the success rate is calculated as the number of successfully completed segments over the total number of segments (Eqn. \ref{eq:success_rate}).  In \LTR, the agent is considered to have failed and the current episode terminates whenever the agent incurs a safety infraction. If the agent fails at a certain segment, it will respawn, stationary at the beginning of the next segment. If the agent successfully completes a segment, it will continue on to the next segment carrying over the current speed. Success rate serves as a proxy for safety, and its relationship with the number of safety infractions is made explicit in Eqn. \ref{eq:sr_safety}. Thus, a higher success rate is better. Success rate is a newly-introduced metric, which intends to improve upon Episode Completion Percentage (ECP) originally defined in \citet{herman2021learn}. An episode here refers to one lap. ECP measures the percentage of a lap the agent successfully completes when spawned once at the starting line. In our experience, ECP has large variance due to the length of an episode / lap. For reference,  a single lap in \texttt{Track01:Thruxton} is 3.8km, whereas CARLA, the \textit{de facto} environment for urban driving research, has in \textit{total} 4.3km of drivable roads in their original benchmark \citep{codevilla2019exploring}. Formally, we define success rate as follows:

\begin{equation}\label{eq:success_rate}
    \text{Success Rate} = \frac{\text{\# Completed Segments}}{\text{Total Number of Segments}}\times 100\%
\end{equation}

\begin{equation}\label{eq:sr_safety}
    \text{Success Rate} + \frac{\text{\# Safety Infractions}}{\text{Total Number of Segments}} = 100\%
\end{equation}

\para{Average Adjusted Track Speed.} Average speed is defined as the total distance traveled over time (Eqn. \ref{eq:average_speed}), which serves as a proxy for performance.
As this is \textit{Formula}-style racing, higher is better; this is the same as Average Adjusted Track Speed (AATS) in Task 1.0 \citep{herman2021learn}.

\begin{equation}\label{eq:average_speed}
    \text{Average Speed} = \frac{\text{Total Distance Traveled}}{\text{Total Time}}
\end{equation}

\para{Total Number of Safety Infractions.} The total number of safety infractions (NSI) is accumulated during the 1-hour ‘practice’ period in Stage 2 of the competition. The agent is considered to have incurred a safety infraction if 2 wheels of the vehicle leave the drivable area, the vehicle collides with an object, or does not make sufficient progress (e.g. get stuck). In Learn-to-Race, the episode terminates upon a safety infraction. A smaller number of safety infractions is better, i.e. the agent is safer. This is a newly-introduced metric to take into consideration that an autonomous agent should remain safe throughout its interaction with the environment \citep{ray2019benchmarking}.

\subsection{Task 2.0 Evaluation Procedure: Cross-domain Model Transfer}
\label{subsection:task2:eval}

Agent assessment is conducted through leaderboard competition, with two distinct stages: \textit{training} and \textit{evaluation}. During the training phase, agents have access to the Thruxton Circuit racetrack (\texttt{Track01:Thruxton}) for model training and the Trac M\^{o}n Anglesey National Circuit racetrack (\texttt{Track02:Anglesey}) for model validation, as well as access to all camera sensors and configurations. During Evaluation, agents will be deployed on the unseen track, North Road at Las Vegas Motor Speedway (\texttt{Track03:Vegas}), and required to maximise the SR and AATS metrics, subject to NSI being as close to zero infractions as possible. The Evaluation stage, itself, is sub-divided into two sub-phases: pre-evaluation and final evaluation. During pre-evaluation, an agent is given a 1-hour `practice period', in which access to all camera sensor types and the execution of model weight updates (i.e., for learning-based approaches) are permitted; in the final evaluation sub-phase, models weights are frozen for inference and access to only RGB cameras is permitted. 

Results of model performance may be provided on the basis of three task execution paradigms---In-domain performance, Cross-domain validation, and Cross-domain test/deployment---against all the metrics in Section \ref{section:metrics}.

\parai{In-domain performance}: \{\texttt{Track01:Thruxton}\}$\rightarrow$\{\texttt{Track01:Thruxton}\} transfer task, where the goal is to maximises an agent's performance on a single track.

\parai{Cross-domain validation}: \{\texttt{Track01:Thruxton}\}$\rightarrow$\{\texttt{Track02:Anglesey}\} transfer task, where the goal is to maximises an agent's cross-domain performance on a validation track.

\parai{Cross-domain deployment (leaderboard submission)}: \{\texttt{Track01:Thruxton}\}$\rightarrow$\{\texttt{Track03:Vegas}\} transfer task, where the goal is to maximises an agent's cross-domain performance on a test track. This option is only available via leaderboard submission, where evaluation on \texttt{Track03:Vegas} is done automatically. Leaderboard and \ltr~Challenge information is provided in Section \ref{sec:challenge}.

\section{\LTR~Autonomous Racing Virtual Challenge} 
\label{sec:challenge}

The first \LTR~Autonomous Racing Virtual Challenge (L2R-ARVC) leaderboard is offered for free as a service to the research community, thanks to the generosity of our sponsors and collaborators. The objective of this instance of the \LTR~challenge is to push the boundary of autonomous technology, with a focus on jointly maximising safety, performance, and generalisability in autonomous driving. We specifically highlight the \ltr~Task 2.0 benchmark (Section \ref{sec:task2}), as we see high-fidelity autonomous racing simulation as a particularly challenging proving ground for autonomous systems--as they are required to adapt to new environments, while making fast decisions, where any safety infraction could have devastating ramifications.

\begin{figure}
    \centering
    \includegraphics[width=0.9\textwidth]{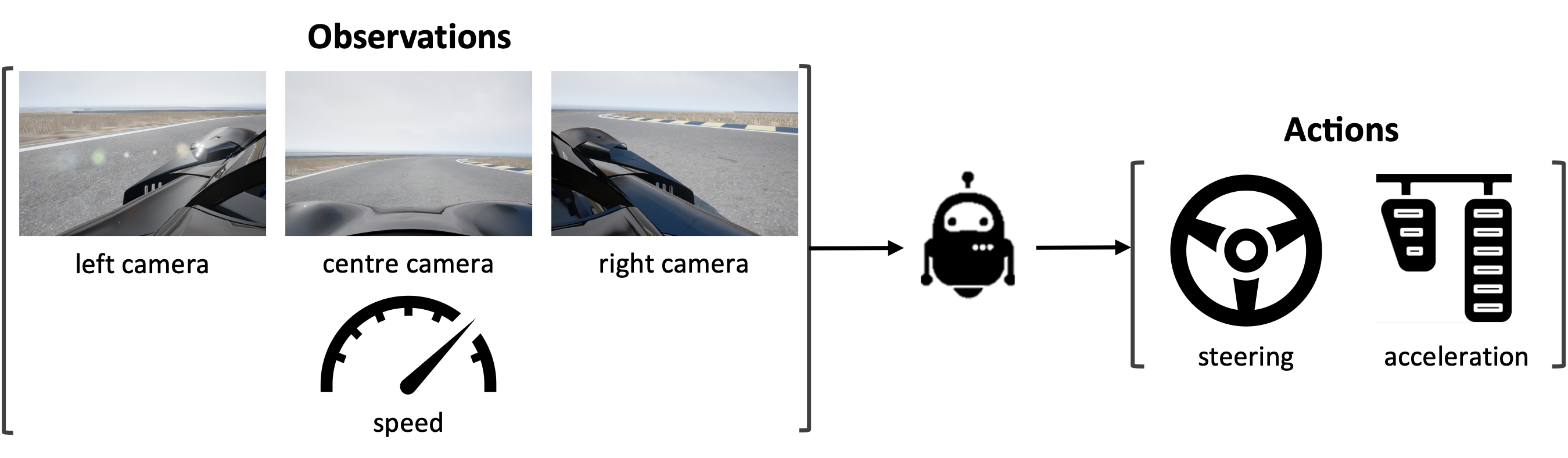}
    \caption{ Agent observations and actions. While the \ltr~provide access to  customisable, multimodal sensory inputs, the agents will ONLY have access to speed and RGB images from cameras placed on the front, right, and left of the vehicle, during evaluation. Based on these observations, the agents determine normalised steering angle and acceleration.  }
    \label{fig:my_label}
\end{figure}

\subsection{Challenge Overview} 

This competition follows from the procedure outlined in the \ltr~Task 2.0 benchmark description (Section \ref{subsection:task2:eval}) and consists of two stages: in Stage 1, participants develop their agents on the Thruxton Circuit track (\texttt{Track01:Thruxton}) to drive as fast as possible, while adhering to the safety constraints; in Stage 2, we test the agents' ability to adapt to a new environment, with a fixed time budget for safe exploration. Participants will be evaluated on an unseen track (\texttt{Track03:Vegas}), with the opportunity to `practice' with unfrozen model weights for a 1-hour period prior to final evaluation. During the practice period, the number of safety infractions will be accumulated as one of the evaluation metrics, under the consideration that an autonomous agent should remain safe throughout its interaction with the environment \citep{ray2019benchmarking}. After the ‘practice’ period, agents are evaluated on the unseen track, against the Task 2.0 benchmark metrics (Section \ref{section:metrics}).

Participants are required to use high-dimensional visual data as inputs. \LTR~provides access to customizable, multimodal sensory inputs. One can access RGB images from any specified location, semantic segmentation, and vehicle states (e.g. pose, velocity). During local development, the participants may use any of these inputs; during final evaluation, the agents will ONLY have access to speed and RGB images from cameras placed on the front, right, and left of the vehicle. 

\subsection{Challenge Infrastructure} 

We have solicited sponsorship from companies such as Amazon Web Services, which enables participants to partake in the competition regardless of the personal compute they have available. The evaluation phase is also supported separately by AWS with 15,000 USD. The competition is hosted on AICrowd, a popular website that enables data science experts and enthusiasts to collaboratively solve real-world problems, through challenges. AICrowd has previously hosted competitions for NeurIPS, OpenAI, Spotify, Uber, Stanford University and UNICEF. Carnegie Mellon University has also generously funded in cash and kind with students helping support the competition. 

Teams are provided with a time budget (currently 300 hours or approximately 20 training runs to convergence) to train their submissions on AWS, and upload for evaluation to AICrowd. Each submission will be evaluated in AWS using a g4dnxlarge/T4 instance. This gives users access to a dedicated node with a modern GPU (Nvidia T4 or better) and CPU. Teams are provided a number of submissions (currently 20 submissions) for a given month. Both budgets are automatically refilled every month. The organisers of the \LTR~Autonomous Racing Virtual Challenge leaderboard reserve the right to assign an additional budget to a team. The organisers also reserve the right to modify the default values of the monthly time budget and/or the number of submissions.

\subsection{Challenge Results}
\label{sec:results}

Official evaluation of Challenge submission is performed in accordance with the \ltr~Task 2.0 benchmark metrics (Section \ref{section:metrics}) and evaluation procedures (Section \ref{subsection:task2:eval}). We further describe Challenge-specific logistics and results.

\para{Challenge stages.} In the Stage 1 (training) phase, participants develop their agents locally, using \texttt{Track01:Thruxton} and optionally \texttt{Track02:Anglesey}, then submit their trained models for evaluation on \texttt{Track01:Thruxton}. The submissions will first be ranked on the SR metric, then submissions with the same SR will be ranked on the basis of AATS. The ten (10) top-ranking teams will be permitted to progress to Stage 2. In Stage 2, the top-ranking participants from Stage 1 will develop cross-domain transfer methods, using \texttt{Track01:Thruxton} and/or \texttt{Track02:Anglesey}; participants will then submit their models for evaluation on the AICrowd server, on the unseen \texttt{Track03:Vegas}. Here, agents are provided with a 1-hour practice period, where access to all camera sensor types and the execution of model weight updates (i.e., for learning-based approaches) are permitted; afterwards, the weights of the approach (at the end of the 1-hour practice phase) are frozen for inference and the the approach is restricted to the use of only RGB camera sensors. Following Stage 2, submissions from the participating teams will first be ranked on success rate, and then submissions with the same SR will be ranked on a weighted sum of the NSI and AATS, based on Eqn. \ref{eq:speed+infractions}, where the subscript $i$ denotes metrics from each participating team:
\begin{equation}\label{eq:speed+infractions}
    \frac{\text{AATS}_i }{\max_i (\text{AATS}_i)} +\max\left(100\%-\frac{\text{NSI}_i}{\text{median}_i(\text{NSI}_i)}, -100\%\right)
\end{equation}

\para{Multiple leaderboards in each stage.} In order to maintain fair comparison, approaches are split across multiple leaderboards, based on how much multimodal/multiview information is used as input: the results from approaches that use single RGB cameras during Stage 2 evaluation are recorded on a different leaderboard form those approaches that use more than one RGB cameras (e.g., RGB-left, RGB-front, and RGB-right) during Stage 2 evaluation. We also provide approach descriptions for the top-ranking method in each category, below.

\para{Leaderboard rankings.} Leaderboard rankings are provided in tables \ref{tab:stage1} and \ref{tab:stage2}; for conciseness, we show SR, AATS, and NSI for each participant/team, across multiple Challenge stages and camera configurations. We also provide approach descriptions of top-ranking participants, below:

\begin{table*}[t]
\caption{Results from the \LTR~Autonomous Racing Virtual Challenge Leaderboard, for the Stage 1 ``\texttt{Track01:Thruxton}$\rightarrow$\texttt{Track01:Thruxton}" in-domain task paradigm.}
\label{tab:stage1}
\scriptsize
\resizebox{\columnwidth}{!}{
\centering
\begin{tabular}{p{0.18\linewidth}p{0.20\linewidth}p{0.20\linewidth}p{0.18\linewidth}p{0.08\linewidth}}
\toprule
\multicolumn{5}{c}{\texttt{Single-camera}} \\
\midrule
\textbf{Participant} & Success Rate (SR; \%)~($\uparrow$) & Speed (AATS; KPH)~($\uparrow$) & Infractions (NSI)~($\downarrow$) & \# Entries \\
\midrule
{saleh9292} & 0.500 & 117.875 & 6.000 & 4 \\
{White-Wolf} & 0.700 & 53.115 & 1.000 & 1 \\
{SS} & 0.700 & 59.953 & 2.000 & 2 \\
{shan\_osphere} & 0.700 & 60.968 & 4.000 & 1 \\
{number9473} & 0.700 & 64.448 & 4.000 & 2 \\
{kire} & 0.800 & 42.943 & 2.000 & 7 \\
{NotSoLate} & 0.900 & 32.485 & 2.000 & 18 \\
{jiangwen\_su} & 0.900 & 57.615 & 1.000 & 20 \\
{agnprz} & 0.900 & 69.045 & 2.000 & 2 \\
{AnimeshSinha1309} & 0.900 & 69.045 & 2.000 & 2 \\
{kobe\_bb} & 0.900 & 78.910 & 2.000 & 4 \\
{boliu0} & 1.000 & 36.140 & 0.000 & 4 \\
{avrl} & 1.000 & 63.080 & 0.000 & 1 \\
{denis9} & 1.000 & 72.000 & 0.000 & 16 \\
{any\_name} & 1.000 & 80.760 & 0.000 & 11 \\
{ling\_thoth} & 1.000 & 93.940 & 0.000 & 6 \\
{TCS\_Autoscape} & 1.000 & 95.960 & 0.000 & 60 \\
{matthew\_howe} & 1.000 & 102.010 & 0.000 & 12 \\
{UniTeam} & 1.000 & 105.350 & 0.000 & 27 \\
{xLab\_UPenn} & 1.000 & 115.660 & 0.000 & 2 \\
{lachlan\_mares} & 1.000 & 126.350 & 0.000 & 15 \\
{Downforce615} & 1.000 & 137.940 & 0.000 & 39 \\
{Werner\_Duvaud} & 1.000 & 152.090 & 0.000 & 15 \\
\bottomrule
\toprule
\multicolumn{5}{c}{\texttt{Multi-camera}} \\
\midrule
\textbf{Participant} & Success Rate (SR; \%)~($\uparrow$) & Speed (AATS; KPH)~($\uparrow$) & Infractions (NSI)~($\downarrow$) & \# Entries \\
\midrule
{Downforce615} & 1.000 & 137.230 & 0.000 & 1 \\
\bottomrule
\end{tabular}
}
\end{table*}

\begin{table*}[t]
\caption{Results from the \LTR~Autonomous Racing Virtual Challenge Leaderboard, for the Stage 2 ``\texttt{Track01:Thruxton}$\rightarrow$\texttt{Track03:Vegas}" cross-domain task paradigm.}
\label{tab:stage2}
\scriptsize
\resizebox{\columnwidth}{!}{
\centering
\begin{tabular}{p{0.18\linewidth}p{0.20\linewidth}p{0.20\linewidth}p{0.18\linewidth}p{0.08\linewidth}}
\toprule
\multicolumn{5}{c}{\texttt{Single-camera}} \\
\midrule
\textbf{Participant} & Success Rate (SR; \%)~($\uparrow$) & Speed (AATS; KPH)~($\uparrow$) & Infractions (NSI)~($\downarrow$) & \# Entries \\
\midrule
{xLab\_UPenn} & 0.000 & 31.098 & 10.333 & 32 \\
{TCS\_Autoscape} & 0.100 & 4.485 & 4.333 & 38 \\
{denis9} & 0.667 & 64.889 & 3.667 & 24 \\
{any\_name} & 1.000 & 30.44 & 0.000 & 6 \\
{Werner\_Duvaud} & 1.000 & 45.253 & 0.000 & 40 \\
{UniTeam} & 1.000 & 73.187 & 0.000 & 45 \\
{matthew\_howe} & 1.000 & 85.22 & 0.000 & 33 \\
{lachlan\_mares} & 1.000 & 92.527 & 0.000 & 34 \\
\bottomrule
\toprule
\multicolumn{5}{c}{\texttt{Multi-camera}} \\
\midrule
\textbf{Participant} & Success Rate (SR; \%)~($\uparrow$) & Speed (AATS; KPH)~($\uparrow$) & Infractions (NSI)~($\downarrow$) & \# Entries \\
\midrule
{UniTeam} & 0.667 & 62.094 & 1.000 & 8 \\
{any\_name} & 1.000 & 51.373 & 0.000 & 3 \\
{lachlan\_mares} & 1.000 & 80.723 & 0.000 & 6 \\
{matthew\_howe} & 1.000 & 84.227 & 0.000 & 1 \\
\bottomrule
\end{tabular}
}
\end{table*}

\begin{itemize}
    \item \textit{Top team on single-camera leaderboard (lachlan\_mares)}: During stage 1 of the competition multiple approaches to solving the problem were evaluated; investigations conducted included: multiple reinforcement learning algorithms (in both discrete and continuous space), track localisation using monte-carlo, Bayesian and machine learning methods and classical vehicle controllers. However, given the constraints of Stage 2, in terms of cross-domain generalisation requirements and given the reduced set of permitted camera sensors, a combination of these methods with the best generalisation capabilities was chosen. The Stage 2 solution used two neural networks to feed information to separate steering and acceleration controllers. The first such model was a semantic segmentation model that required low inference time and high accuracy; the architecture chosen was a custom implementation consisting of an EfficientNet-V2-Small encoder, paired with a Feature Pyramid Network (FPN) decoder. The segmentation model was trained using 384x512 augmented greyscale images, rather than RGB, to reduce the domain gap between the Thruxton, AngleseyNational, and VegasNorthRoad tracks. The classifier was a fully-connected network, which took a latent vector from the segmentation model and classified the track into one of three zone types. The zone types permitted the use of multiple parameter settings for the steering and acceleration controllers. The steering and velocity controllers both rely on predictions from the segmentation model. An estimate of the track centerline and track curvature was calculated using the boundaries of the predicted road surface: the zone classifier allowed relaxed parameters in high-speed regimes, so that the vehicle could exceed speeds permitted by only relying on the camera's field of view.
    \item \textit{Top team on multi-camera leaderboard (matthew\_howe)}: We provide a baseline that can, in a single lap, safely adapt to an unseen race-track and achieve competitive lap times. Our solution consists of a perception module which informs a model predictive controller. The perception module is responsible for processing observations in the form of a camera feed into a representation of the local limits of the road. These road limits and additional safety constraints are used by the model predictive controller to provide steering and acceleration inputs. \textit{Perception}---Using ground truth segmentation masks provided by the simulation on the Thruxton circuit, a Feature Pyramid Network (FPN) \citep{lin2017feature} with an EfficentNet \citep{tan2019efficientnet} backbone was trained to classify whether a given image pixel belongs to the drive-able portion of the race track or not. Generalisation across racetracks was achieved via data augmentation. Specifically a combination of horizontal flipping and brightness changes were found to enable acceptable inference performance on the Anglesey racetrack. The segmentation mask boundaries are projected onto the ground plane using camera intrinsic and extrinsic calibrations to yield local track boundaries. To smooth the often jagged mask boundaries, this module uses cubic polynomial fitting to produce the final track limits used downstream. \textit{Control}---A model predictive controller \citep{schneuing2020multipurposempc} takes observed track limits and plans a path that both stays on the track and achieve high speeds. Given a set of physical constraints, a spatial bicycle model is used to predict achievable values for yaw and velocity. The controller plans a series of future states that are within track limits, gets to the last point on the trajectory as quickly as possible, do no exceed maximum lateral acceleration and be able to slow down to a minimum cornering speed by the end of the observed trajectory. Both the lateral acceleration limit and cornering speed were tuned manually by observing vehicle behaviour on the training circuits. A lateral acceleration limit of 2.0m/s2 was found to rarely exceed grip limits and a minimum cornering speed of 10m/s ensured that excess velocity was not carried into corners. The lateral acceleration limit serves to reduce cornering speed; leading to less opportunity for loss of control. Enforcing cornering speeds prevent the vehicle entering turns too dangerously, such as diving deep and exceeding track limits. Combining these hand-crafted constraints led to more reliable and consistent cornering behaviour.
\end{itemize}

\section{Insights and Future Directions}\label{sec:future}

\para{Towards safe and performant reinforcement learning.} We envision racing to be a particularly challenging task for safe learning algorithms due to the inherent tension between speed and safety.  In this competition, we hope to encourage the use of  safe RL on multimodal perception for autonomous racing, where autonomous agents need to identify and avoid unsafe scenarios under the complex vehicle dynamics, and make sub-second decisions in a fast-changing environment. While the top teams in Stage 1 can successfully complete the lap without incurring safety infractions, there is still significant performance gap with human, as the speed record at Thruxton Circuit is 237 km/h in comparison to the average speed of 152km/h achieved by the best-performing team (see Table \ref{tab:stage1}). 
This thrust focuses on the extension, development, and benchmarking of new safe learning algorithms for autonomous racing.

\para{Towards generalisation in reinforcement learning.} We also envision the competition to be an opportunity to test generalisability of RL agents.
Perhaps not surprisingly, the top-performing teams used the modular pipeline, where the control module depends on MPC and/or hand-crafted heuristics, instead of end-to-end RL to achieve cross-domain generalisation. This points to the fact that training generalist RL agents that can quickly adapts to new environments/tasks remains an open research challenge \citep{laskin2021urlb, ghosh2021generalization}.

\para{Towards low-latency and generalisable visual feature-extraction.} Autonomous systems need to characterise and acquire semantic understanding of their environment, in order to identify unsafe scenarios and avoid obstacles. In this challenge, agents are endowed with visual observations of their environment, from which they must extract meaningful features (e.g., distances to road boundaries, other agents, and obstacles), while remaining invariant to irrelevant features (e.g., shadows, glare, sky, vegetation). To add to the challenge, the vision encoder is subject to stringent requirements on inference time, due to the nature of the autonomous racing task. High latency can adversely impact agent performance, where, as discussed in prior art \citep{strobel2020accurate}, perception stacks in autonomous race-cars account for nearly 60\% of total latency. Moreover, even though we want to maximise performance on the tracks that the agent directly interacts with, we do not want to overfit to the training tracks; the visual encoder must generalise to unseen contexts. The top-performing teams use low-latency EfficientNet \citep{tan2019efficientnet} as backbones, and improve generalisation via data augmentation, e.g., greyscale and horizontal flipping. This thrust focuses on the further development of new visual encoding strategies for autonomous racing.

\para{Distributed training and optimisation.} The optimisation of an RL agent depends on the comprehensiveness of its exploration, during interaction with its environment. Prior art pursued distributed RL architectures, to maximise environment interactions of some performance oriented policy. In the context of safe RL for autonomous racing \citep{chen2021safe}, we have \textit{two} decoupled policies to optimise (i.e., a safety policy and a performance policy), yielding multiple options and hierarchies for distributed training. This development thrust focuses on implementing and experimenting with different distributed reinforcement learning training paradigms, e.g.: multiple model instances + single environment, single model instance + multiple environments, centralised versus decentralised replay memory, centralised versus decentralised safety backup policies/critics/experts, heterogeneous types of policies (on-policy, off-policy) and safety critics (cost-limit regressor, rules-based expert, learnable safety critic), etc \citep{liang2018rllib}.

\para{Towards effective transfer imitation learning.} Recent work in urban driving settings couple imitation learning (IL) objectives with driving policies that have knowledge of the underlying action prior distribution. We hypothesise that this distribution-awareness would provide agents with robustness to noise artifacts in the training data, would provide a window into agents' intentions for more interpretable predictions, would yield improved unseen generalisation, and would help bypass common issues in imitation- and transfer learning such as causal confusion \citep{de2019causal} and negative transfer \citep{parisotto2015actor}. An interesting direction for future work is in the selection of more informative priors, e.g., those that incorporate logical rules for appropriate multi-agent interaction on the racetrack. Another important direction for future work is in characterising the causal relationships between observations, actions, and rewards in a scene. This causal structure can be represented \textit{explicitly}, by way of causal knowledge graphs (enabling explainability and counterfactual reasoning), or it can be represented \textit{implicitly}, by way of learning identifiable latent representations. Regardless of explicit or implicit representation, the goal would be to eliminate extraneous connections/edges (disturbances, confounders) in the underlying causal structure, thereby reducing causal confusion when transferring an observation model from a source domain to a target domain.


\para{Planned extensions of the \ltr~platform:}
\begin{itemize}
    \item \parai{Towards simulation-to-real transfer.} In designing the Stage 2 of the competition, we intended the transfer between racetracks as a proxy for the ultimate goal of sim-to-real transfer. The \LTR~framework will interface with the real-world vehicle software stacks used in the Roborace Challenge and Indy Autonomous Challenge competitions, via the Robot Operating System (ROS). Our goal is to support flexible transfer of algorithms---from simulation in \ltr~to real-world autonomous racing vehicles.
    \item \parai{Future framework extensions.} We believe that a primary direction for future tasks must be towards increased simulation complexity and capabilities. The \LTR~framework will feature new evaluation procedures and test-cases that match upcoming Roborace competition scenarios, such as virtual static/dynamics obstacles in Roborace Metaverse and multi-agent settings, across many new tracks; we will introduce new tasks and challenges, accordingly.
    \item \parai{Future outreach and modes of engagement with \ltr.} The \LTR~team wishes to continue outreach with the scientific community, through further publications, challenges, and collaborations. We also organise and sponsor workshops at notable scientific publication venues, such as the \href{https://learn-to-race.org/workshop-ai4ad-ijcai2022/}{\textit{Workshop on Artificial Intelligence for Autonomous Driving}} (AI4AD; co-located with the \textit{International Joint Conference on Artificial Intelligence}, IJCAI 2022, in Vienna) and the \href{https://learn-to-race.org/workshop-sl4ad-icml2022/}{\textit{Workshop on Safe Learning for Autonomous Driving}} (SL4AD; co-located with the \textit{International Conference on Machine Learning}, ICML 2022, in Baltimore).
\end{itemize}

\section{Conclusions}\label{sec:conclusions}

In this paper, we summarised the results of the inaugural \LTR~Autonomous Racing Virtual Challenge, which encouraged interdisciplinary research and development towards safe, performant, generalisable autonomous systems. We introduce the \ltr~Task 2.0 benchmark, with new metrics and task definitions. Then we provide an overview of the deployment, evaluation, and rankings, which used the new \ltr~Task 2.0 benchmark and received over 20,100 views, 437 participants, 46 teams, and 733 model submissions---from 88+ unique institutions, in 58+ countries. 

\acks{There are several entities that the authors would like to thank --- principally, the \textit{Masters in Computational Data Science} (MCDS) program, in the School of Computer Science, at Carnegie Mellon University: MCDS provided student development support, via capstone projects that were proposed and led by JF, SK, SGa, AK, and EN. We thank AICrowd, Amazon AWS, and CMU students for extensive challenge sponsorship and support: Rabiul Islam, Sharada Mohanty, Cameron Peroc, Joe Fontaine, and Shravya Bhat. We also thank the Challenge participants themselves for their engagement and feedback---in particular, the \textit{lachlan\_mares} and \textit{matthew\_howe} participants for contributing their approach descriptions. This work was supported, in part, by a doctoral research fellowship from Bosch Research. The views expressed in this article do not necessarily represent those of the aforementioned entities. JF is the corresponding author, reachable at \textit{jmf1@alumni.cmu.edu}.
}

\clearpage

\bibliography{main}
\bibliographystyle{apalike}

\appendix

\end{document}